\documentclass[11pt]{article}

\usepackage[preprint]{acl}

\usepackage{times}
\usepackage{latexsym}

\usepackage[T1]{fontenc}

\usepackage[utf8]{inputenc}

\usepackage{microtype}

\usepackage{inconsolata}

\usepackage{graphicx}

\usepackage{amsmath}
\usepackage{amssymb}
\usepackage{mathtools}
\usepackage{amsthm}
\usepackage{subfigure}

\title{Toward LLMs Beyond English-Centric Development}

\author{Sho Takase \hspace{1.5em} Ukyo Honda \\
  CyberAgent \\
  \texttt{\{honda\_ukyo, takase\_sho\}@cyberagent.co.jp} \\
 }

\begin{document}
\maketitle
\begin{abstract}
Through an analysis of sequences generated by open-weight large language models (LLMs), we demonstrate that LLMs are heavily biased toward English.
While continual pre-training is commonly used to adapt LLMs to a target language, we show that it does not offer a cost advantage over training from scratch, even for improving cultural understanding in the target language.
These findings suggest that dedicated per-language investment may become increasingly important for future LLM development, rather than relying primarily on the expansion of English-centric resources.
\end{abstract}

\section{Introduction}
\label{sec:introduction}
Since the initial report on the remarkable performance of the large language model (LLM)~\cite{NEURIPS2020_1457c0d6}, numerous groups have introduced LLMs including the Llama, Qwen, and Gemma series~\cite{touvron2023llama1,bai2023qwentechnicalreport,gemmateam2024gemmaopenmodelsbased}.
It appears that novel LLMs yielding performance improvements are being reported almost on a daily basis.
However, because the majority of existing LLMs focus on English and major benchmark datasets are predominantly in English, comparatively little attention has been devoted to discussions involving other languages.

In fact, some studies have reported that they trained LLMs on datasets whose significant portions consist of English text~\cite{touvron2023llama2,bai2023qwentechnicalreport}.
For example, the Llama 2 pre-training corpus contains approximately 90\% English documents.
Because the large-scale Web crawled corpus, CommonCrawl, includes only about 40\% English documents\footnote{\href{https://commoncrawl.github.io/cc-crawl-statistics/plots/languages}{https://commoncrawl.github.io/cc-crawl-statistics/plots/languages}}, their pre-training data is heavily biased toward English compared with the natural distribution of Web documents.
In this paper, we estimate the language distribution of training data for each LLM based on its generated sequences, and we demonstrate that an LLM is strongly biased toward English even when it is designed as a multilingual model (Figure \ref{fig:lang_distribution}).

To efficiently obtain an LLM that achieves superior performance in a specific language, previous studies have conducted continual pre-training~\cite{fujii2024continual}.
We analyze such continual pre-training models and show that continual pre-training does not offer a cost advantage over training from scratch for developing LLMs specialized for a particular language.
We show that continual pre-training can improve performance in the target language, but the improvement is commensurate with the additional computational cost incurred during continual pre-training.
In particular, we demonstrate that performance on cultural understanding related to a specific language is proportional to the logarithmic scale of the training cost for that language, rather than to the total training cost (Figure \ref{fig:score_on_training_cost}).
These results confirm that the efficiency of continual pre-training for such cultural understanding tasks is comparable to that of training from scratch.
Based on these findings, we discuss future directions for LLM development.

\section{Language Distribution of LLMs}
\label{sec:lang_dist}
\begin{figure*}[!t]
  \centering 
    \includegraphics[width=14cm]{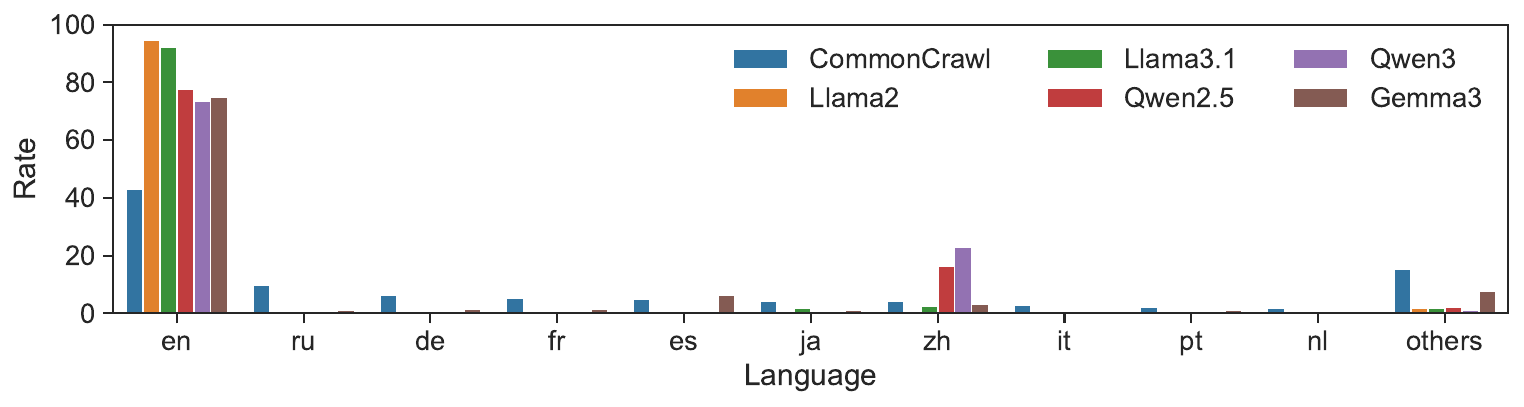}
   \caption{Estimated language distribution of pre-training data for each LLM based on randomly generated sequences. This figure focuses on top 10 languages in CommonCrawl.}
   \label{fig:lang_distribution}
\end{figure*}

\begin{table*}[!t]
  \centering{}
  \footnotesize
  \begin{tabular}{ l | c c c c c c c c c c c | c} \hline
  Model & En & Ru & De & Fr & Es & Ja & Zh & Th & Te & Bn & Sw & Average \\ \hline \hline
  Qwen2.5-32B-Instruct & 87.2 & 72.8 & 71.2 & 64.4 & 86.4 & 76.0 & 80.8 & 69.2 & 58.0 & 82.4 & 56.8 & 73.2 \\
  Qwen3-32B & 95.6 & 86.4 & 88.0 & 82.4 & 91.6 & 77.6 & 89.6 & 88.8 & 84.0 & 83.2 & 74.4 & 85.6 \\
  Llama3.1-8B-Instruct & 78.8 & 62.4 & 67.2 & 64.0 & 73.6 & 53.2 & 60.0 & 60.8 & 53.2 & 54.0 & 54.8 & 62.0 \\
  Gemma-3-27B-it & 96.0 & 86.0 & 86.8 & 80.4 & 92.4 & 75.6 & 84.8 & 88.8 & 82.8 & 84.8 & 84.4 & 85.7 \\ \hline
  \end{tabular}
  \caption{6-shot performance of each LLM on the MGSM test set. We take model names from HuggingFace Hub.\label{tab:exp_mgsm_all_lang}}
\end{table*}

Although some groups report the construction process of their LLMs in technical reports~\cite{touvron2023llama2, qwen2025qwen25technicalreport, gemmateam2025gemma3technicalreport}, these reports do not reveal detailed statistics about their pre-training datasets.
In this paper, we estimate the language distribution of pre-training datasets based on the assumption that the sequences generated by each LLM strongly correlate with its training data.
In fact, previous studies have shown that we can extract training data from LLMs~\cite{274574,nasr2025scalable,xu2025magpie}.

\paragraph{Estimation of Distribution}
We ask each LLM to generate 100,000 sequences, each consisting of 1,000 tokens, from the BOS token.
We then identify the language of each sequence using fastText~\cite{joulin2016fasttextzipcompressingtextclassification, grave-etal-2018-learning}.
Finally, we regard the resulting language distribution of the generated sequences as an estimate of the language distribution of the pre-training data.
To evaluate the reliability of this process, we compute the Kendall rank correlation coefficient~\cite{kendall1938measure} between the reported language distribution and our estimated one for Llama 2~\cite{touvron2023llama2}, which is one of the few studies that reveal the language distribution of their pre-training corpus.
As a result, we obtained coefficients of 0.77 and 0.80 for Llama 2 7B and 13B, respectively.
Because these values are significantly high, we believe that our estimation method is reasonable\footnote{We also investigate LLM-jp and OLMo that do not reveal their language distribution but make their pre-training corpus publicly available. Appendix \ref{appendix:correlation_on_others} provides further details.}.

\paragraph{Comparison of Distribution}
Figure \ref{fig:lang_distribution} shows the estimated language distributions of widely used open-weight LLMs and CommonCrawl.
For CommonCrawl, we sampled 100,000 documents containing more than 1,000 characters and identified their languages using fastText for a fair comparison to LLMs.
For the LLMs, we focus on pre-training models trained from scratch by each group.
Appendix \ref{appendix:detail_lang_dist} provides the details of distributions.

Figure \ref{fig:lang_distribution} shows that the proportion of English in the pre-training datasets is much higher than that in the CommonCrawl corpus.
In other words, the LLMs in Figure \ref{fig:lang_distribution} are strongly biased toward English.
Thus, although these LLMs achieve good performance in English, they may not reach sufficient quality in other languages.
In fact, such LLMs exhibit a substantial drop in performance relative to English when evaluated on language-agnostic benchmark datasets that assess language-independent abilities, such as solving math problems, across various languages.
Table \ref{tab:exp_mgsm_all_lang} presents the few-shot performance of each LLM on the multilingual grade school math (MGSM) dataset~\cite{shi2023language}, which contains math problems translated from English into multiple languages.
We set the number of shots to 6, following the setting used in the previous study~\cite{shi2023language}.
For example, as shown in Table \ref{tab:exp_mgsm_all_lang}, the performance in Russian is about 10 points lower than in English, even though Russian is the second most frequent language in CommonCrawl.

\section{Effectiveness and Efficiency of Continual Pre-training}
\label{sec:evaluate_continual_pretrain}
To adapt an LLM to a target language, previous studies have applied continual pre-training with the target language~\cite{yang2023bigtranslateaugmentinglargelanguage,10.1007/978-3-031-45392-2_15,zhao2024llamaenglishempiricalstudy,cui2024efficienteffectivetextencoding,fujii2024continual}.
These studies reported that continual pre-training can produce a strong LLM in the target language at a substantially lower computational cost than training from scratch.
In this section, we re-evaluate the performance of continual pre-training on both language-agnostic benchmark datasets and benchmark datasets that require cultural understanding of the target language to answer the questions.

 \begin{table*}[!t]
  \centering{}
  \footnotesize
  \begin{tabular}{ l | c c c c | c c c } \hline
  & \multicolumn{4}{c|}{Language-agnostic} & \multicolumn{3}{c}{} \\
        & \multicolumn{2}{c}{MGSM} & \multicolumn{2}{c|}{MMLU ProX} & \multicolumn{3}{c}{Japanese cultural understanding} \\ 
  Model & En & Ja & En & Ja & NIILCQA & Part of JMMLU & JAQKET \\ \hline \hline
  \multicolumn{8}{c}{Continual pre-trained models and their base model} \\ \hline \hline
  Llama-3.1-8B-Instruct & 78.8 & 53.2 & 43.5 & 28.7 & 30.7 & 71.6 & 27.5 \\
  Llama-3.1-Swallow-8B-Instruct-v0.5 & 76.8 & 60.0 & 43.0 & 36.2 & \textbf{58.3} & 85.7 & 65.5 \\ \hline
  Qwen2.5-32B & 92.0 & 82.4 & 61.1 & 51.0 & 35.4 & 84.3 & 48.4 \\
  Qwen2.5-bakeneko-32B-Instruct & \textbf{94.4} & \textbf{84.0} & \textbf{71.9} & \textbf{64.9} & 32.3 & 83.5 & 51.2 \\ \hline
  Qwen2.5-32B-Instruct & 87.2 & 76.0 & 68.7 & 61.2 & 26.8 & 84.3 & 49.9 \\
  ABEJA-Qwen2.5-32b-Japanese-v1.0 & 90.8 & 32.8 & 66.3 & 58.1 & 42.5 & \textbf{90.9} & \textbf{70.5} \\
  ELYZA-Shortcut-1.0-Qwen-32B & 61.6 & 69.2 & 61.7 & 62.9 & 36.2 & 87.3 & 61.7 \\ \hline \hline
  \multicolumn{8}{c}{LLMs with pre-training data containing a large amount of Japanese text} \\ \hline \hline
  Sarashina2.2-3B-Instruct-v0.1 & 76.0 & 60.4 & 36.1 & 30.2 & 52.0 & 81.9 & 68.2 \\
  LLM-jp-3.1-13B-Instruct4 & 57.2 & 66.0 & 29.8 & 28.5 & 51.2 & 80.8 & 69.4 \\ \hline
  \end{tabular}
  \caption{Performance on language-agnostic benchmarks and Japanese cultural understanding benchmarks. We use 6-shot for MGSM, 5-shot for MMLU Pro\-X, 4-shot for NIILCQA and JMMLU, and zero-shot for JAQKET.\label{tab:exp_japanese_bench}}
\end{table*}

\paragraph{Model}
We focus on continual pre-training models adapted to Japanese because there are various models.
Specifically, we evaluate Llama-3.1-Swallow-8B-Instruct-v0.5 trained from Llama-3.1-8B-Instruct~\cite{grattafiori2024llama3herdmodels}, Qwen2.5-bakeneko-32B-Instruct trained from Qwen2.5-32B~\cite{qwen2025qwen25technicalreport}, and ABEJA-Qwen2.5-32B-Japanese-v1.0 and ELYZA-Shortcut-1.0-Qwen-32B trained from Qwen2.5-32B-Instruct~\cite{qwen2025qwen25technicalreport}.
These models are instruction-tuned after continual pre-training.

\paragraph{Dataset}
For language-agnostic benchmark tasks, we focus on math problem and massive multilingual language understanding (MMLU) datasets.
We use the Japanese portions of MGSM~\cite{shi2023language} and MMLU Pro\-X~\cite{xuan-etal-2025-mmlu}, which are translated from English grade school math problems~\cite{cobbe2021trainingverifierssolvemath} and MMLU Pro~\cite{NEURIPS2024_ad236edc}, respectively.
We also report the English portions of these datasets as a reference.
In addition, we use the NIILC question answering dataset (NIILCQA)~\cite{NIILCQA}, the subset of JMMLU related to Japanese culture~\cite{yin-etal-2024-respect}, and JAQKET\footnote{\href{https://github.com/kumapo/JAQKET-dataset}{github.com/kumapo/JAQKET-dataset}} as benchmark datasets to evaluate cultural understanding of Japan.
We use the LM Evalution Harness\footnote{\href{https://github.com/EleutherAI/lm-evaluation-harness}{github.com/EleutherAI/lm-evaluation-harness}} for language-agnostic benchmarks and FlexEval\footnote{\href{https://github.com/sbintuitions/flexeval/}{github.com/sbintuitions/flexeval/}} for Japanese cultural understanding benchmarks.

\begin{figure*}[!t]
\centering
    \includegraphics[width=15cm]{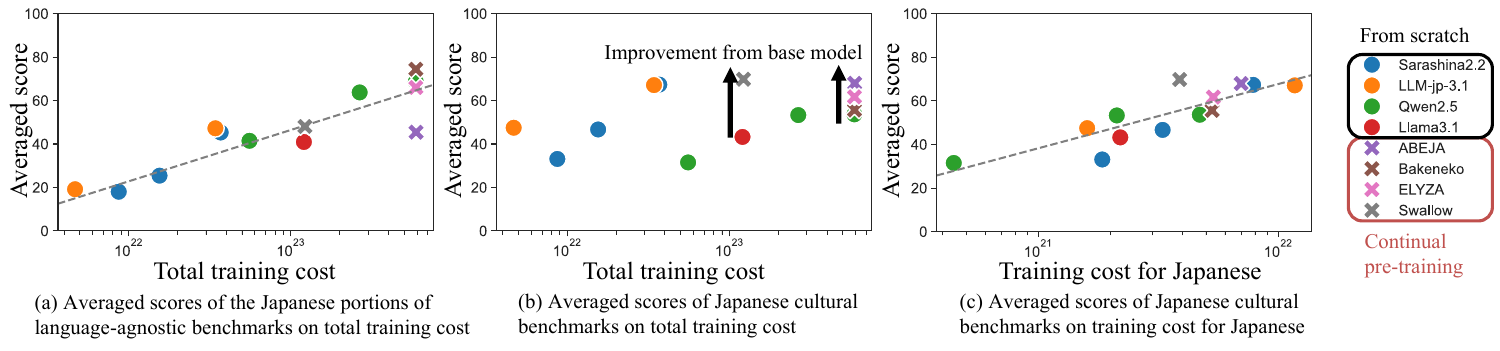}
    \caption{Averaged benchmark scores for each training cost. The vertical arrows in (b) indicate the performance improvement from the base model through continual pre-training.}
    \label{fig:score_on_training_cost}
\end{figure*}

 \paragraph{Result 1: Effectiveness}

Table \ref{tab:exp_japanese_bench} shows the results of each LLM on language-agnostic benchmarks and Japanese cultural understanding benchmarks.
This table also reports the performance of LLMs whose pre-training data contains a large amount of Japanese text, for reference.
For the language-agnostic benchmarks, Llama-3.1-Swallow-8B-Instruct-v0.5 improves performance over its base model in Japanese portions.
Qwen2.5-bakeneko-32B-Instruct also improves the performance.
Thus, continual pre-training in the target language can improve performance on language-agnostic benchmarks.
However, as shown in the results of ABEJA-Qwen2.5-32B-Japanese-v1.0 and ELYZA-Shortcut-1.0-Qwen-32B, continual pre-training may also degrade performance in the target language.
Therefore, it is important to design an appropriate training strategy carefully when adapting a strong model.
Moreover, for Japanese cultural understanding benchmarks, all models except Qwen2.5-bakeneko-32B-Instruct improve upon the performance of their base models.

\paragraph{Result 2: Efficiency}
Table \ref{tab:exp_japanese_bench} demonstrates that continual pre-training mostly improves performance in the target language.
However, how efficient is continual pre-training?
In other words, does continual pre-training provide a larger performance gain per unit of computational cost compared to pre-training from scratch?
To investigate this question, we illustrate the relationship between performance and computational cost.
Specifically, we compute the average score of the Japanese portions of the language-agnostic benchmarks and Japanese cultural benchmarks.
We then examine the relationship among these averaged scores, the total computational cost, and the computational cost spent on training with Japanese texts.
We use the product of the parameter size and the number of consumed tokens as the measure of computational cost.
To compute the amount of Japanese training data, we use the estimated ratio obtained in Section \ref{sec:lang_dist}.

Figure \ref{fig:score_on_training_cost} shows the relationship between performance and training cost.
To construct this figure, we additionally evaluated models with varying parameter sizes for Qwen2.5, Sarashina2.2, and LLM-jp-3.1.
We also plot a regression line based on the points.
As shown in Figure \ref{fig:score_on_training_cost} (a), the scores on the Japanese portions of language-agnostic benchmarks improve linearly with respect to the logarithmic scale of the total training cost.
This result implies that continual pre-training with the target language does not have a substantial impact on the performance of language-agnostic benchmarks for the target language.

On the other hand, Figure \ref{fig:score_on_training_cost} (b) shows that continual pre-training models achieve substantial improvements on the Japanese cultural understanding benchmarks when evaluated in terms of the total training cost.
Based on this figure, continual pre-training appears to be a remarkably cost-efficient approach when considering only the total training cost.
In contrast, when we focus on the training cost specifically spent on Japanese, Figure \ref{fig:score_on_training_cost} (c) indicates that improvements are proportional to the logarithmic scale of the training cost for Japanese.
Thus, the gains obtained through continual pre-training arise only from the additional training cost allocated to Japanese.
In other words, the cost-performance trend of continual pre-training follows approximately the same scaling as training from scratch for the target language.

\section{Discussion}
\label{sec:discussion}
Based on the findings from Sections \ref{sec:lang_dist} and \ref{sec:evaluate_continual_pretrain}, performance on language-agnostic benchmarks is likely to continue improving under recent English-centric LLM development, as such performance primarily depends on the total training cost.
Furthermore, previous studies have shown that LLMs can improve performance on language-agnostic benchmarks by prompting the model to think in the predominant language, i.e., English, rather than by explicitly learning each language~\cite{shi2023language}.
As a result, there is limited incentive to incorporate large amounts of non-English data into LLM training when the objective is solely to improve performance on language-agnostic benchmarks.
Consequently, LLMs tend to exhibit weaker cultural understanding for languages that are not major in their training corpora.

Moreover, it is worth questioning whether language-agnostic benchmarks are truly language-independent.
In common benchmark construction pipelines, reasoning task datasets are first created in English and then translated into multiple languages.
Thus, these benchmarks are implicitly influenced by English (or Western-centric) concepts.
In fact, \citet{singh-etal-2025-global} reported that the MMLU dataset contains questions biased toward Western cultural perspectives, such as \textit{US History} and \textit{US Law}.

As LLMs are becoming foundational tools for a wide range of applications, it is undesirable for them to embed strong cultural biases.
To better understand and mitigate such biases, we emphasize the importance of developing diverse culture-specific benchmarks in addition to language-agnostic benchmarks.
At a minimum, our findings encourage consideration beyond improvements on language-agnostic benchmarks alone.

\section{Conclusion}
In this study, we show that the pre-training datasets of open-weight LLMs contain a disproportionately large amount of English text compared with the language distribution of the large-scale Web crawled corpus, CommonCrawl.
In addition, we indicate that continual pre-training is approximately as cost-efficient as training from scratch, even for improving cultural understanding in a target language.
We hope that our findings help advance LLM development beyond English-centric paradigms.

\section*{Limitations}
\paragraph{Language Distribution Estimation}
To estimate the language distribution of pre-training corpora, we analyze sequences generated by each LLM.
To assess the reliability of this approach, we compare the estimated language distribution with the reported distribution for Llama 2~\cite{touvron2023llama2}, which is one of the few studies that disclose the language distribution of their training corpus.
We observe a significantly high correlation between the two distributions.
In addition, as described in Appendix \ref{appendix:correlation_on_others}, we also investigate the correlation for LLM-jp and OLMo models and observe similarly strong correlations.
We also confirm that our estimation is stable across different random seeds.
However, we acknowledge that it would be desirable to validate this correlation for other models if such information were available.

Furthermore, we also investigate fine-tuned versions of Llama 2 7B and 13B.
Their Kendall rank correlation coefficients are 0.62 and 0.60, respectively, which are lower than those of the corresponding pre-training models.
Therefore, we focus on pre-training models when estimating language distributions.

\paragraph{Target Language}
In this study, we focus on Japanese as the target language for analyzing continual pre-training, because a wide range of continual pre-training models and cultural understanding benchmark datasets are available for Japanese.
To the best of our knowledge, no other language currently offers a comparable combination of both resources for a systematic analysis.

Since the procedure of continual pre-training is not language-specific, we believe that the findings obtained from Japanese continual pre-training models can be applied to models for other languages.
This is analogous to how neural scaling laws, which were primarily established through experiments on English corpora~\cite{kaplan2020scalinglawsneurallanguage,NEURIPS2022_c1e2faff}, are widely accepted as applicable across languages. However, we acknowledge that investigating continual pre-training models for additional languages would further strengthen the generality of our findings.

\paragraph{Configuration of Continual Pre-training}
In this study, our analysis is limited to models with up to 32B parameters.
Because pre-training LLMs at larger scale requires substantial computational resources, even when using continual pre-training, publicly available LLMs at substantially larger scales are limited.
Although we anticipate similar trends for larger models, verifying whether Figure \ref{fig:score_on_training_cost} (c) generalizes to larger parameter scales remains an important direction for future work.

We note that the $R^2$ values of the regression lines in Figure~\ref{fig:score_on_training_cost} (a) and (c) are 0.60 and 0.57, respectively. Since our analysis relies on diverse models independently developed rather than a controlled training strategy and uses several datasets for evaluation, such variation is expected~\cite{ivgi-etal-2022-scaling}.

In addition, some recent LLMs adopt Mixture-of-Experts (MoE) architecture~\cite{shazeer2017} but we focus on dense architecture in this study.
While it is an important direction to investigate their properties such as neural scaling laws~\cite{liew2025scaling}, MoE models have a large number of total parameters, and thus, they require substantial computational resources for training, similar to the discussion on model size above.
As a result, publicly available MoE models that have undergone continual pre-training are scarce, and we leave this investigation for future work.

\paragraph{Cultural Understanding Benchmarks}
Recent studies have introduced benchmarks for each language~\cite{isbarov-etal-2025-tumlu,koto-etal-2024-arabicmmlu,li-etal-2024-cmmlu,koto-etal-2023-large,togmanov-etal-2025-kazmmlu,yin-etal-2024-respect}, highlighting the importance of evaluating LLMs beyond English-centric perspectives.
Our study complements these efforts by providing quantitative evidence of the language distribution in training data, offering an empirical basis for systematically discussing the efficacy of LLMs across languages, rather than investigating internal representations~\cite{zhong-etal-2025-language}.
Our analysis is limited to text-based modality, and extending this investigation to multimodal models, such as constructing benchmarks for VLMs~\cite{vayani2024alm,mogrovejo2024cvqa}, is beyond the scope of this study.

\bibliography{reference}
\bibliographystyle{acl_natbib}

\clearpage

\appendix

\section{Additional Validation of Language Distribution Estimation}
\label{appendix:correlation_on_others}

To examine the reasonableness of our language distribution estimation method based on sequences generated by each LLM, we investigate the Kendall rank correlation coefficient for LLM-jp and OLMo models~\cite{llmjp2024llmjpcrossorganizationalprojectresearch,groeneveld-etal-2024-olmo}, in addition to Llama 2.
Unlike Llama 2, these models do not disclose the language distribution of their pre-training corpus but they do provide the corpus itself.
Therefore, we sample data from the released corpus and compute the correlation coefficient between the language distribution of the sampled corpus and that of the sequences generated by those models.

We focus on LLM-jp-3-1.8B, LLM-jp-3-13B and OLMo-1B.
For LLM-jp models, LLM-jp Corpus v3\footnote{\href{https://gitlab.llm-jp.nii.ac.jp/datasets/llm-jp-corpus-v3}{https://gitlab.llm-jp.nii.ac.jp/datasets/llm-jp-corpus-v3}} was used for pre-training.
Because this corpus is divided into multiple domains, we sample documents from each domain in proportion to the reported number of tokens in that domain.
We then concatenate the sampled documents and further sample 100,000 documents from the concatenated set to estimate the language distribution.
For OLMo, Dolma was used for pre-training~\cite{soldaini-etal-2024-dolma}, and thus, we sample 100,000 documents from it.
The resulting Kendall rank correlation coefficients are 0.61 for LLM-jp-3-1.8B, 0.63 for LLM-jp-3-13B, and 0.71 for OLMo-1B.
The slightly lower coefficients for LLM-jp models may be attributed to the approximate nature of our corpus sampling, as the pre-training corpus is divided into multiple domains and we estimate the overall distribution through proportional sampling from each domain.
Since these values indicate strong correlations, we consider our estimation method to be reasonable.

In addition, we examine the robustness of our estimation method with respect to random seeds during inference.
We generate sequences from Llama 2 13B using two different random seeds.
The resulting Kendall rank correlation coefficients are 0.70 and 0.71, which are similarly high.
This result indicates that our estimation is stable across different random seeds.

\section{Hyper-parameters during Sequence Generation}
\label{appendix:hyper_parameters_sequence_generation}

 \begin{table}[!t]
  \centering{}
  \footnotesize
  \begin{tabular}{ c | r } \hline
  Hyper-parameter & Value \\ \hline \hline
  Temperature & 1.0 \\
  Top\_p & 1.0 \\
  Top\_k & -1 \\
  max\_tokens & 1,000 \\
  stop & None \\
  ignore\_eos & True \\
  repetition\_penalty & 1.0 \\ \hline
  \end{tabular}
  \caption{Hyper-parameters used in sequence generation.\label{tab:hyper_parameters_generation}}
\end{table}

To generate sequences randomly in Section \ref{sec:lang_dist}, we use vLLM~\cite{kwon2023efficient} with its default hyper-parameters.
Table \ref{tab:hyper_parameters_generation} lists the corresponding values.

\section{Details of Language Distribution}
\label{appendix:detail_lang_dist}

 \begin{table*}[!t]
  \centering{}
  \footnotesize
  \begin{tabular}{ l | r r r r r r r r r r r } \hline
  Model & \multicolumn{1}{c}{En} & \multicolumn{1}{c}{Ru} & \multicolumn{1}{c}{De} & \multicolumn{1}{c}{Fr}
  & \multicolumn{1}{c}{Es} & \multicolumn{1}{c}{Ja} & \multicolumn{1}{c}{Zh} & \multicolumn{1}{c}{It} & \multicolumn{1}{c}{Pt} & \multicolumn{1}{c}{Nl} & \multicolumn{1}{c}{others} \\ \hline
  CommonCrawl & 43.0 & 9.8 & 6.4 & 5.1 & 5.1 & 4.2 & 4.1 & 2.8 & 2.3 & 1.9 & 15.3 \\
  Llama 2 & 89.7 & 0.2 & 0.2 & 0.2 & 0.1 & 0.1 & 0.1 & 0.1 & 0.1 & 0.1 & 9.1 \\
  Llama 3.1 & 92.3 & 0.4 & 0.3 & 0.2 & 0.3 & 1.8 & 2.6 & 0.1 & 0.1 & 0.1 & 1.7 \\
  Qwen 2.5 & 77.7 & 0.8 & 0.6 & 0.8 & 0.5 & 0.8 & 16.2 & 0.3 & 0.3 & 0.1 & 1.9 \\
  Qwen 3 & 73.4 & 0.6 & 0.3 & 0.3 & 0.6 & 0.4 & 22.8 & 0.1 & 0.5 & 0.0 & 1.0 \\
  Gemma 3 & 74.9 & 1.3 & 1.4 & 1.6 & 6.2 & 1.1 & 3.1 & 0.9 & 1.2 & 0.8 & 7.5 \\
  LLM-jp-3 & 63.4 & 0.0 & 0.0 & 0.0 & 0.0 & 36.4 & 0.1 & 0.0 & 0.0 & 0.0 & 0.0 \\
  Sarashina2.2 & 73.7 & 0.5 & 0.3 & 0.4 & 0.4 & 21.3 & 2.0 & 0.1 & 0.2 & 0.0 & 1.1 \\ \hline
  \end{tabular}
  \caption{Details of language distribution for CommonCrawl and each LLM. We represent the percentage of each language.\label{tab:detail_lang_dist}}
\end{table*}

Table \ref{tab:detail_lang_dist} shows the estimated language distribution from sampled CommonCrawl corpus and sequences generated by each LLM except for Llama 2.
For Llama 2, we used the language distribution reported in the original paper~\cite{touvron2023llama2}.
For this estimation, we used Llama-3.1-8B\footnote{\href{https://huggingface.co/meta-llama/Llama-3.1-8B}{huggingface.co/meta-llama/Llama-3.1-8B}}, Qwen2.5-3B\footnote{\href{https://huggingface.co/Qwen/Qwen2.5-3B}{huggingface.co/Qwen/Qwen2.5-3B}}, Qwen3-4B-Base\footnote{\href{https://huggingface.co/Qwen/Qwen3-4B-Base}{huggingface.co/Qwen/Qwen3-4B-Base}}, Gemma-3-4B-pt\footnote{\href{https://huggingface.co/google/gemma-3-4b-pt}{huggingface.co/google/gemma-3-4b-pt}}, LLM-jp-3-1.8B\footnote{\href{https://huggingface.co/llm-jp/llm-jp-3-1.8b}{huggingface.co/llm-jp/llm-jp-3-1.8b}}, and Sarashina2.2-3B\footnote{\href{https://huggingface.co/sbintuitions/sarashina2.2-3b}{huggingface.co/sbintuitions/sarashina2.2-3b}}.
We note that LLM-jp-3-1.8B is the pre-training model of LLM-jp-3.1-1.8B-Instruct4.

\section{Amount of Japanese Training Data}
\label{appendix:amount_ja_train_data}

Table \ref{tab:amount_ja_train_data} shows the estimated number of Japanese tokens trained by base model and the number of Japanese tokens during continual pre-training for each model.

\begin{table*}[!t]
  \centering{}
  \scriptsize
  \begin{tabular}{ l | c c } \hline
  Model & Estimated Japanese tokens in pre-training & Japanese tokens in continual pre-training \\ \hline
  Llama3.1-Swallow-8B-Instruct-v0.5\footnote{\href{https://huggingface.co/tokyotech-llm/Llama-3.1-Swallow-8B-Instruct-v0.5}{huggingface.co/tokyotech-llm/Llama-3.1-Swallow-8B-Instruct-v0.5}} & 274B & 210B \\ 
  Qwen2.5-Bakeneko-32B-Instruct-v2\footnote{\href{https://huggingface.co/rinna/qwen2.5-bakeneko-32b-instruct}{huggingface.co/rinna/qwen2.5-bakeneko-32b-instruct}} & 144B & 18B \\
  ABEJA-Qwen2.5-32B-Japanese-v1.0\footnote{\href{https://huggingface.co/abeja/ABEJA-Qwen2.5-32b-Japanese-v1.0}{huggingface.co/abeja/ABEJA-Qwen2.5-32b-Japanese-v1.0}} & 144B & 70B \\
  ELYZA-Shortcut-1.0-Qwen-32B\footnote{\href{https://huggingface.co/elyza/ELYZA-Shortcut-1.0-Qwen-32B}{huggingface.co/elyza/ELYZA-Shortcut-1.0-Qwen-32B}} & 144B & 20B \\ \hline \hline
  \multicolumn{3}{c}{LLMs with pre-training data containing a large amount of Japanese text} \\ \hline \hline
  Sarashina2.2-3B-Instruct-v0.1\footnote{\href{https://huggingface.co/sbintuitions/sarashina2.2-3b-instruct-v0.1}{https://huggingface.co/sbintuitions/sarashina2.2-3b-instruct-v0.1}} & 2339B & N/A \\
  LLM-jp-3.1-13B-Instruct4\footnote{\href{https://huggingface.co/llm-jp/llm-jp-3.1-13b-instruct4}{https://huggingface.co/llm-jp/llm-jp-3.1-13b-instruct4}} & 911B & N/A \\ \hline
  \end{tabular}
  \caption{The estimated number of Japanese tokens in base model training and the number of Japanese tokens during continual pre-training.\label{tab:amount_ja_train_data}}
\end{table*}

\section{MGSM Performance with Thinking in Each Language}
\label{appendix:six_shot_performance_mgsm}
\begin{table*}[!t]
  \centering{}
  \scriptsize
  \begin{tabular}{ l | c c c c c c c c c c c | c} \hline
  Model & En & Ru & De & Fr & Es & Ja & Zh & Th & Te & Bn & Sw & Average \\ \hline \hline
  \multicolumn{13}{c}{Thinking in English} \\ \hline \hline
  Qwen2.5-32B-Instruct & 87.2 & 79.6 & 77.2 & 74.8 & 83.6 & 71.2 & 76.8 & 77.6 & 61.6 & 80.8 & 52.0 & 74.8 \\
  Qwen3-32B & 95.6 & 90.0 & 88.4 & 88.4 & 90.4 & 87.2 & 87.6 & 89.6 & 82.8 & 85.6 & 79.2 & 87.7 \\
  Llama3.1-8B-Instruct & 80.0 & 68.4 & 72.0 & 71.2 & 75.6 & 58.8 & 66.4 & 67.6 & 54.0 & 58.4 & 56.0 & 66.2 \\
  Gemma-3-27B-it & 96.0 & 91.2 & 88.0 & 85.6 & 92.0 & 81.2 & 89.2 & 89.6 & 84.8 & 88.0 & 83.6 & 88.1 \\
  Llama3.1-Swallow-8B-Instruct-v0.5 & 76.8 & 64.8 & 63.6 & 68.4 & 68.8 & 66.4 & 62.8 & 50.4 & 36.0 & 45.2 & 41.6 & 58.6 \\
  Qwen2.5-Bakeneko-32B-Instruct-v2 & 94.4 & 90.8 & 90.4 & 85.2 & 87.2 & 85.6 & 88.4 & 89.6 & 64.0 & 83.6 & 60.8 & 83.6 \\
  ABEJA-Qwen2.5-32B-Japanese-v1.0 & 90.8 & 89.6 & 84.4 & 83.2 & 86.4 & 83.2 & 84.0 & 86.4 & 59.2 & 81.2 & 51.6 & 80.0 \\
  ELYZA-Shortcut-1.0-Qwen-32B & 61.6 & 63.6 & 50.4 & 60.8 & 86.0 & 80.0 & 63.6 & 66.0 & 55.2 & 68.0 & 44.0 & 63.6 \\
  Sarashina2.2-3B-Instruct-v0.1 & 76.0 & 56.0 & 62.8 & 62.0 & 66.8 & 68.4 & 60.4 & 22.4 & 7.2 & 17.6 & 8.4 & 46.2 \\
  LLM-jp-3.1-13B-Instruct4 & 57.2 & 55.6 & 52.8 & 54.0 & 57.6 & 64.0 & 50.8 & 28.4 & 19.2 & 30.0 & 11.6 & 43.7 \\ \hline \hline
  \multicolumn{13}{c}{Thinking in Chinese} \\ \hline \hline
  Qwen2.5-32B-Instruct & 85.6 & 86.4 & 83.6 & 82.0 & 86.0 & 76.4 & 80.8 & 86.8 & 54.0 & 76.0 & 62.0 & 78.1 \\
  Qwen3-32B & 94.4 & 88.4 & 88.4 & 84.8 & 91.6 & 80.8 & 89.6 & 88.4 & 85.2 & 85.6 & 73.2 & 86.4 \\
  Llama3.1-8B-Instruct & 74.8 & 66.4 & 65.2 & 53.6 & 70.4 & 54.4 & 60.0 & 17.6 & 38.0 & 48.0 & 55.2 & 54.9 \\
  Gemma-3-27B-it & 92.0 & 87.2 & 85.2 & 84.4 & 90.0 & 78.4 & 84.8 & 86.4 & 80.4 & 82.0 & 82.4 & 84.8 \\
  Llama3.1-Swallow-8B-Instruct-v0.5 & 62.4 & 50.8 & 54.8 & 52.0 & 62.4 & 53.6 & 55.6 & 40.8 & 26.4 & 26.8 & 34.0 & 47.2 \\
  Qwen2.5-Bakeneko-32B-Instruct-v2 & 94.0 & 90.8 & 89.6 & 84.4 & 90.8 & 83.2 & 87.2 & 88.4 & 62.4 & 86.0 & 62.0 & 83.5 \\
  ABEJA-Qwen2.5-32B-Japanese-v1.0 & 88.0 & 85.2 & 86.0 & 86.4 & 86.8 & 80.4 & 84.0 & 57.2 & 62.4 & 44.0 & 55.2 & 74.1 \\
  ELYZA-Shortcut-1.0-Qwen-32B & 73.6 & 70.8 & 60.8 & 69.6 & 84.8 & 79.6 & 82.0 & 68.0 & 56.4 & 63.2 & 59.2 & 69.8 \\
  Sarashina2.2-3B-Instruct-v0.1 & 65.6 & 47.6 & 46.4 & 51.6 & 52.8 & 60.8 & 54.0 & 24.8 & 5.6 & 13.6 & 8.0 & 39.2 \\
  LLM-jp-3.1-13B-Instruct4 & 61.2 & 52.0 & 55.2 & 53.2 & 49.2 & 61.6 & 16.4 & 26.0 & 20.4 & 22.8 & 6.8 & 38.6 \\ \hline \hline
  \multicolumn{13}{c}{Thinking in Japanese} \\ \hline \hline
  Qwen2.5-32B-Instruct & 78.4 & 74.0 & 75.2 & 62.4 & 79.2 & 76.0 & 76.4 & 58.4 & 57.6 & 64.8 & 54.8 & 68.8 \\
  Qwen3-32B & 96.4 & 85.2 & 90.0 & 86.8 & 92.0 & 77.6 & 87.2 & 74.8 & 83.2 & 84.0 & 72.4 & 84.5 \\
  Llama3.1-8B-Instruct & 57.6 & 62.0 & 58.0 & 43.2 & 58.4 & 53.2 & 50.8 & 41.2 & 42.0 & 44.0 & 48.4 & 50.8 \\
  Gemma-3-27B-it & 90.8 & 85.6 & 85.2 & 81.2 & 86.0 & 75.6 & 82.8 & 83.6 & 79.6 & 81.6 & 81.2 & 83.0 \\
  Llama3.1-Swallow-8B-Instruct-v0.5 & 68.8 & 62.4 & 63.2 & 55.2 & 61.2 & 60.0 & 51.2 & 43.6 & 33.2 & 46.4 & 38.0 & 53.0 \\
  Qwen2.5-Bakeneko-32B-Instruct-v2 & 91.2 & 86.4 & 84.4 & 83.2 & 86.4 & 84.0 & 82.4 & 87.6 & 60.8 & 82.8 & 60.4 & 80.9 \\
  ABEJA-Qwen2.5-32B-Japanese-v1.0 & 39.2 & 60.4 & 67.2 & 45.6 & 71.2 & 32.8 & 50.4 & 37.2 & 28.4 & 34.4 & 44.4 & 46.5 \\
  ELYZA-Shortcut-1.0-Qwen-32B & 74.4 & 59.2 & 69.2 & 63.2 & 75.2 & 69.2 & 34.4 & 55.6 & 54.4 & 64.4 & 45.2 & 60.4 \\
  Sarashina2.2-3B-Instruct-v0.1 & 66.0 & 55.2 & 50.0 & 57.2 & 54.8 & 60.4 & 58.8 & 21.2 & 8.8 & 15.2 & 10.8 & 41.7 \\
  LLM-jp-3.1-13B-Instruct4 & 77.6 & 56.4 & 57.6 & 64.4 & 62.8 & 66.0 & 56.0 & 35.2 & 22.8 & 35.2 & 7.6 & 49.2 \\ \hline
  \end{tabular}
  \caption{6-shot performance with thinking in each language on the MGSM test set.\label{tab:exp_mgsm_in_thinking_each_lang}}
\end{table*}

Table \ref{tab:exp_mgsm_in_thinking_each_lang} shows the few-shot performance of each LLM on the MGSM test set when we prompt the models to think in English, Chinese, or Japanese.
Following \citet{shi2023language}, we provide each few-shot example as a pair consisting of a question in the original language and an answer in the target language.
For example, in the case of Russian with thinking in English, the few-shot example consists of a Russian question and its corresponding English answer.

Table \ref{tab:exp_mgsm_in_thinking_each_lang} indicates that Llama3.1-8B-Instruct and Gemma-3-27B-it achieve the best performance when thinking in English, consistent with the findings reported by \citet{shi2023language}.
In contrast, the performance of Qwen2.5-32B-Instruct and Qwen3-32B when thinking in Chinese is comparable to or better than their performance when thinking in English.
In addition, LLM-jp-3.1-13B-Instruct4 achieves the best performance when thinking in Japanese.
These results indicate that an LLM can achieve superior performance when thinking in a language that is frequent in its pre-training corpus.
In other words, strong performance when prompting an LLM to think in English suggests that English occupies a large proportion of its pre-training corpus.
\end{document}